\documentclass[10pt,twocolumn,letterpaper]{article}

\usepackage{cvpr}              %

\usepackage[accsupp]{axessibility} %

\makeatletter
\apptocmd\@maketitle{{\teaserfigure{}\par}}{}{}
\makeatother

\def\myshift#1{\raisebox{0.5ex}}
\newcommand{\teasernobox}{
\begin{subfigure}[b]{\linewidth}
    \centering
            \includegraphics[width=\linewidth]{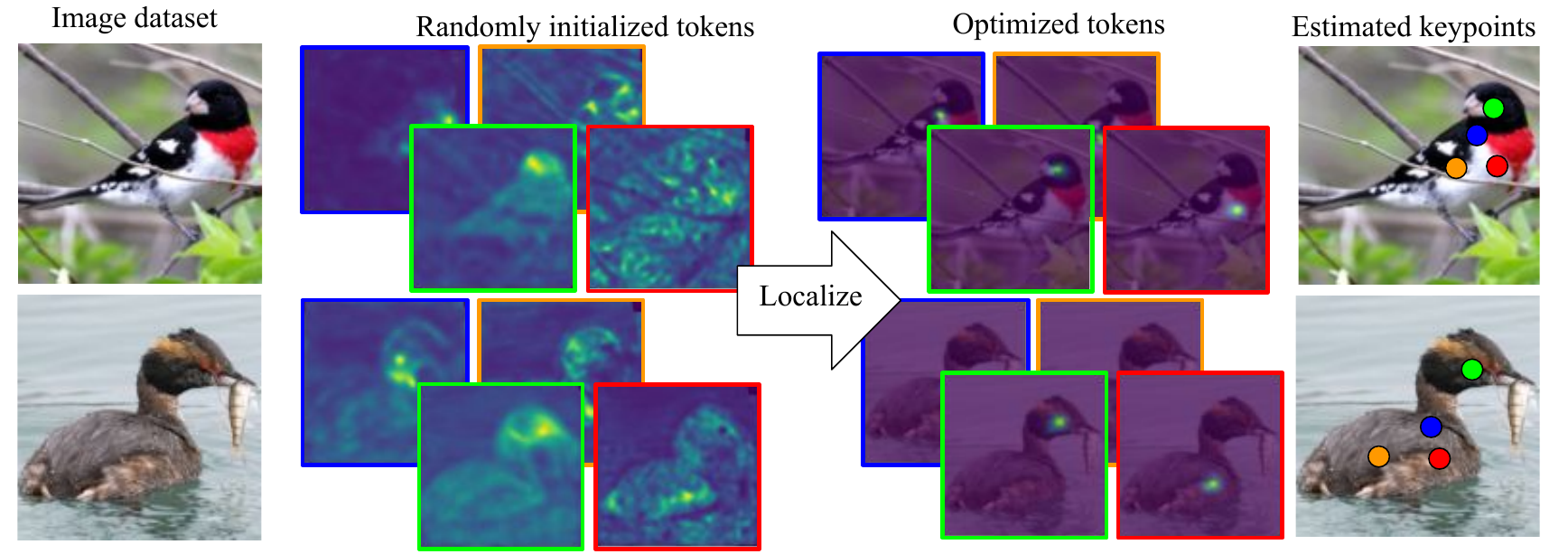}
\end{subfigure}
}

\newcommand{\teaserfigure}{
\vspace{-8mm}
\captionsetup{type=figure}

\teasernobox

\vspace{-2mm}
\setcounter{figure}{0} %
\captionsetup{type=figure}
\captionof{figure}{
{\bf Teaser} --
we propose an unsupervised method to learn keypoints based on optimizing text embeddings of latent diffusion models~\cite{stableDiffusion}.
Our method is motivated by the fact that random text tokens already respond roughly consistently to semantically similar regions.
By promoting localization 
we obtain unsupervised keypoints that outperform the state-of-the-art. 
}
\vspace{4mm}
\label{fig:teaser}
}
\vspace{-1em}

\usepackage{algorithm}
\usepackage{algpseudocode}
\usepackage{graphicx} %
\usepackage{multicol} %
\usepackage{booktabs}
\usepackage{multirow}
\usepackage{amsmath}
\usepackage{siunitx}
\usepackage{amsfonts}
\usepackage{bm}
\usepackage{cancel}
\usepackage{soul}
\usepackage{makecell}
\usepackage{balance}

\usepackage[dvipsnames]{xcolor}

\usepackage{wrapfig}
\usepackage{xspace}

\DeclareMathOperator*{\argmax}{arg\,max}

\def \customparskip {.5em}
\renewcommand{\paragraph}[1]{\vspace{\customparskip}\noindent\textbf{#1}.}

\usepackage{enumitem}
\setlist[itemize]{noitemsep,leftmargin=*,topsep=.5em}
\setlist[enumerate]{noitemsep,leftmargin=*,topsep=.5em}

\definecolor{turquoise}{cmyk}{0.65,0,0.1,0.3}
\definecolor{purple}{rgb}{0.65,0,0.65}
\definecolor{dark_green}{rgb}{0, 0.5, 0}
\definecolor{orange}{rgb}{0.9, 0.6, 0.1}
\definecolor{red}{rgb}{0.8, 0.2, 0.2}
\definecolor{darkred}{rgb}{0.6, 0.1, 0.05}
\definecolor{blueish}{rgb}{0.0, 0.3, .6}
\definecolor{light_gray}{rgb}{0.7, 0.7, .7}
\definecolor{pink}{rgb}{1, 0, 1}
\definecolor{greyblue}{rgb}{0.25, 0.25, 1}

\newcommand{\comment}[1]{}

\newcommand{\image}{\mathbf{X}}

\newcommand{\transform}{\mathcal{T}}
\newcommand{\numTokens}{N}

\newcommand{\iToken}{n}

\newcommand{\latent}{\mathbf{z}}

\newcommand{\target}{\mathbf{G}}
\newcommand{\numkeypoints}{K}
\newcommand{\fpsNumSamples}{\kappa}
\newcommand{\gaussianSort}{\mathcal{N}}

\newcommand{\prompt}{\mathbf{y}}
\newcommand{\textenc}{\tau}
\newcommand{\embedding}{\mathbf{e}}
\newcommand{\params}{{\boldsymbol{\theta}}}
\newcommand{\noise}{\epsilon}
\newcommand{\loss}[1]{\mathcal{L}_{\text{#1}}}
\newcommand{\calN}{\mathcal{N}}
\newcommand{\query}{\mathbf{Q}}
\newcommand{\key}{\mathbf{K}}
\newcommand{\layer}{l}
\newcommand{\IE}{\mathbb{E}}
\newcommand{\IR}{\mathbb{R}}

\newcommand{\attentionMap}{\mathbf{M}}

\definecolor{cvprblue}{rgb}{0.21,0.49,0.74}
\usepackage[pagebackref,breaklinks,colorlinks,citecolor=cvprblue]{hyperref}

\def\paperID{2583} %
\def\confName{CVPR}
\def\confYear{2024}

\title{Unsupervised Keypoints from Pretrained Diffusion Models}

\author{
  Eric Hedlin\textsuperscript{1}\quad
  Gopal Sharma\textsuperscript{1}\quad
  Shweta Mahajan\textsuperscript{1, 2}\quad
  Xingzhe He\textsuperscript{1}\quad
  Hossam Isack\textsuperscript{3}\\
  Abhishek Kar\textsuperscript{3}\quad
  Helge Rhodin\textsuperscript{1}\quad
  Andrea Tagliasacchi\textsuperscript{4, 5, 6}\quad
  Kwang Moo Yi\textsuperscript{1}
  \\
  \textsuperscript{1} University of British Columbia\quad
  \textsuperscript{2} Vector Institute for AI\quad
  \textsuperscript{3} Google Research \\
  \textsuperscript{4} Google DeepMind\quad 
  \textsuperscript{5} Simon Fraser University\quad
  \textsuperscript{6} University of Toronto
  \\
  {\small \url{https://stablekeypoints.github.io/}}
}

\begin{document}
\maketitle
\begin{abstract}
Unsupervised learning of keypoints and landmarks has seen significant progress with the help of modern neural network architectures, but performance is yet to match the supervised counterpart, making their practicability questionable.
We leverage the emergent knowledge within text-to-image diffusion models, towards more robust unsupervised keypoints.
Our core idea is to find text embeddings that would cause the generative model to consistently attend to compact regions in images (i.e. keypoints).
To do so, we simply optimize the text embedding such that the cross-attention maps within the denoising network are localized as Gaussians with small standard deviations.
We validate our performance on multiple datasets: the CelebA, CUB-200-2011, Tai-Chi-HD, DeepFashion, and Human3.6m datasets.
We achieve significantly improved accuracy, sometimes even outperforming supervised ones, particularly for data that is non-aligned and less curated.
Our code is publicly available at the \href{https://stablekeypoints.github.io/}{project page}.
\end{abstract}
\begin{figure*}[t]
\centering
\includegraphics[width=\textwidth]{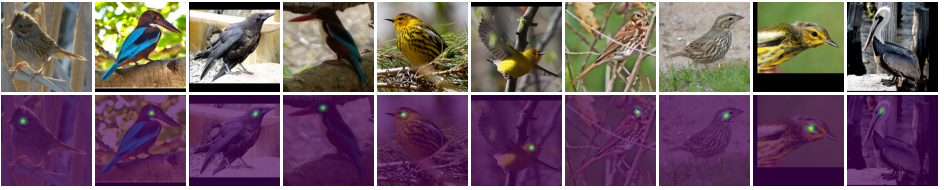}
\caption{
{\bf Example attention maps --}
we show example attention maps for a selected learned keypoint for the CUB-200-2011 dataset, on the CUB-aligned subset.
As shown, our keypoint attention map responds consistently across varying images.
}
\label{fig:attn_maps}
\end{figure*}

\section{Introduction}
\label{sec:intro}

Keypoints or landmarks have played a critical role in computer vision for various task including image matching~\cite{SIFT}, 3D reconstruction \cite{jabberi202368}, and motion tracking \cite{luiten2023dynamic, wang2023tracking}.
Similarly to many other areas of computer vision, research has quickly adopted supervised learning to tackle this problem~\cite{openpose, liu2022recent}.
However, labeling is tedious and sometimes even ambiguous---for example, it is difficult to consistently decide which keypoints on a human face are the ``most important''.
Researchers have therefore been investigating unsupervised approaches~\cite{thewlis2017unsupervised, zhang2018unsupervised, lorenz2019unsupervised, jakab2018unsupervised, latentkeypointgan, autolink}.
These are typically implemented as autoencoders paired with hand-crafted intermediate layers, or losses that enforce spatial locality and equivariance of keypoint locations under deformation.
However, as we will show later, these methods struggle with non-preprocessed data, and their performance is heavily reliant on knowing the ground truth location of objects, clearly limiting their practical applicability.

To enhance the learning of unsupervised keypoints, we draw inspiration from the demonstrated success of scaling up datasets~\cite{sun2017revisiting}.
For example, in natural language processing performance has recently improved to a great extent thanks to large models and data~\cite{llama, palm, gpt4}.
Similarly, in computer vision, the performance of text-to-image models~\cite{stableDiffusion, imagen, dalle} has drastically improved thanks to the availability of \textit{extra} large datasets~\cite{laion5b}.
However, unsupervised keypoint learning typically assumes class-specific datasets, \eg, animals that have a shared skeleton that connects keypoints, and these datasets are small in scale.

Rather than collecting larger domain-specific datasets, we instead propose to leverage the knowledge stored within large generative models, such as Stable Diffusion~\cite{stableDiffusion}.
This has been shown to be very effective across a number tasks~\cite{TaleOFTwoFeats, diffusionHyperfeatures, unsupervised_correspondences_using_sd, emergentCorrespondences, chen2022diffusiondet,
slime, odise, baranchuk2021label, diffuseAttendSegment, DiffuMask, DiffSegmenter, textToMask, azizi2023synthetic, clark2023text, dreamFusion, latentNerf, magic3D}, but, to the best of our knowledge, it has not yet found application for the task of keypoint learning.
Our main idea is to localize ``important'' keypoints by finding text embeddings that \textit{consistently} correspond to a distinct location in images of a certain object class.
This idea is rooted in the observation that, even with random text embeddings, the attention maps for various images roughly correspond to regions that are semantically similar; see \cref{fig:teaser}. 
Therefore, text embeddings carry semantic meaning, which could be used to relate collections of images to each other; see \cref{fig:attn_maps}.

We find embeddings that are specific to certain locations by enforcing localized attention maps. 
In more detail, we propose to find (\ie, optimize) a set of tokens in a text embedding that locally responds in the Stable Diffusion cross-attention layers.
We enforce locality by maximizing the similarity of the attention responses of each token to a single-mode Gaussian distribution.
Thanks to the way the cross-attention layers are constructed within Stable Diffusion, this simple objective also prevents the different tokens from attending to the same locations in an image, a common degenerate solution that typically requires explicit workarounds~\cite{tusk}.

We evaluate our method on established benchmarks: CelebA~\cite{celeba}, CUB-200-2011~\cite{cub}, Tai-Chi-HD~\cite{taichi}, DeepFashion~\cite{DeepFashion}, and Human3.6m~\cite{human36m}.
Our approach yields results on par with state-of-the-art methods for well-curated and aligned datasets, while notably enhancing performance for in-the-wild setups, particularly with unaligned data, sometimes even surpassing fully supervised baselines.
\section{Related Work}
Below we review the literature of finding keypoints in unsupervised and supervised fashion, along with work that exploits large pre-trained models like stable-diffusion for lower-level computer vision tasks.

\paragraph{Learning keypoints with supervision}
Pose estimation and landmark estimation are fundamental problems in computer vision.
They naturally arise in various tasks, including human~\cite{zheng2023deep} and animal pose estimation~\cite{jiang2022animal}, hand~\cite{chen2020survey} and face landmark estimation~\cite{wu2019facial}, and object pose tracking~\cite{marullo20236d}. 
Many fully supervised methods find different ways to induce some prior within the model to better capture the task at hand, such as using part affinity fields~\cite{openpose}, temporal consistency for video data~\cite{russello2022t}, spacial relationships~\cite{xu2022zoomnas}, and geometry constraints~\cite{jau2020deep} among others.
While fully supervised methods have excelled in categories with abundant labeled data, such as human pose estimation, their major drawback is the insatiable need for large and high-quality datasets~\cite{nandy2022audacity, jiang2022animal, qu2022towards}. 
The scalability of gathering such extensive and meticulously annotated data for every conceivable object category remains a significant drawback~\cite{nandy2022audacity, jiang2022animal, qu2022towards}.

\paragraph{Learning keypoints via self-supervision}
The amount of unlabeled data far exceeds that of labeled data, so unsupervised keypoint estimation methods attempt to take advantage of this. 
Self-supervised keypoint detection often relies on tracking how keypoints move with image changes and uses various constraints for known transformations~\cite{jakab2018unsupervised, thewlis2017unsupervised, lorenz2019unsupervised, Siarohin_2019_CVPR, zhang2018unsupervised, scoops}, but these methods can struggle with background modeling~\cite{Siarohin_2019_CVPR, zhang2018unsupervised} and pose variations~\cite{scoops}. 
One can also rely on image reconstruction to learn keypoints.
Some methods use GANs to generate images from keypoints~\cite{latentkeypointgan, ganseg}, but this often results in training instability.
Alternatively, auto-encoders can be also used~\cite{autolink, zhang2018unsupervised}, but these require training from scratch on each dataset.
Our method neither suffers from GAN training instability, nor requires dataset fine-tuning.
Finally, there exist self-supervised methods that exploit skeletal representations~\cite{jakab2020self, papandreou2018personlab, autolink}. 
However, many of these approaches generally require known keypoint connectivity and video data \cite{jakab2020self, papandreou2018personlab}, and often face limitations in background handling and generalizability to objects within the same class~\cite{jakab2020self, papandreou2018personlab, autolink}.
Our method has no object-specific priors, and generalizes well due to the large dataset used by the pre-trained diffusion models.

\paragraph{Diffusion models for image understanding}
Recently, large image diffusion models have reached impressive image generation quality~\cite{imagen, dalle, stableDiffusion}. 
These models learn priors for real images within the latent space of the diffusion model, and provide a useful initialization for many down-stream tasks such as image correspondence~\cite{TaleOFTwoFeats, diffusionHyperfeatures, unsupervised_correspondences_using_sd, emergentCorrespondences}, object detection~\cite{chen2022diffusiondet}, 
semantic segmentation ~\cite{slime, odise, baranchuk2021label, diffuseAttendSegment, DiffuMask, DiffSegmenter, textToMask}, and image classification~\cite{azizi2023synthetic, clark2023text}.
Interestingly, without requiring any retraining, these models demonstrate an innate ability to understand 3D spatial configurations~\cite{dreamFusion, latentNerf, magic3D}.
Recent work in each of these areas has shown the emergent power of these large models, most of them using the model without any modifications or extra supervision required.
More relevant to our work, \citet{prompt-to-prompt} found that the pre-trained Stable Diffusion~\cite{stableDiffusion} model's cross-attention maps connect text tokens to semantically relevant areas in images.

\paragraph{Correspondences via diffusion models}
Among works that re-purpose diffusion models, of high relevance, is the effectiveness of diffusion models in correspondence estimation tasks~\cite{TaleOFTwoFeats, diffusionHyperfeatures, unsupervised_correspondences_using_sd, emergentCorrespondences}.
\citet{unsupervised_correspondences_using_sd} optimizes the attention map for a specific point in a source image and finds the corresponding activation in a target image.
However, this method requires a \textit{query} to be provided in the source image.
While our method shares the same inspiration of utilizing attention maps, critically,
rather than optimizing the embedding given a single image, we optimize an embedding given a dataset of images from a given object class.
In other words, our method discovers on its own, where to focus, rather than relying on user input.
Our task is therefore changed from image matching between \textit{two} images, to semantic matching across \textit{all} images within the dataset.

\begin{figure*}[t]
  \centering
  \includegraphics[width=\linewidth]{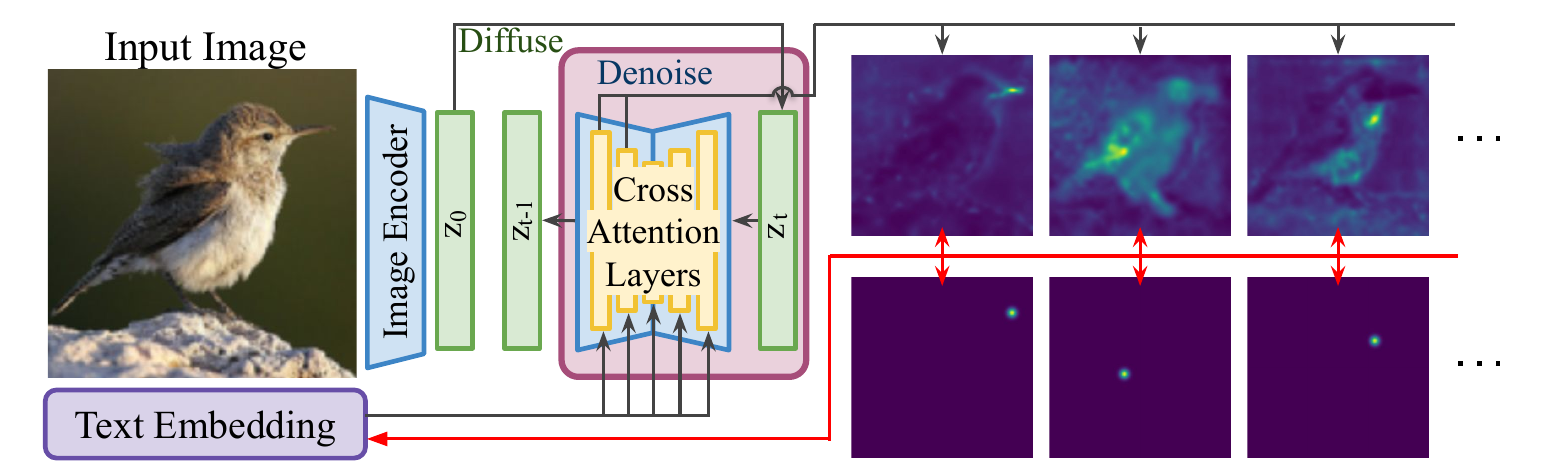}
  \caption{
  \textbf{Overview --} 
  we pass a randomly initialized text embedding into Stable Diffusion~\cite{stableDiffusion} and extract the attention maps.
  We then optimize the text embedding to have localized attention maps, by supervising them to become a single-mode Gaussian distribution, drawn at the location of their maxima.
  We also enforce attention maps to be transformation equivariant to small affine transformations on images.
  We repeat this process over a set of training images, which after optimization provides a set of $\numkeypoints$ keypoints.
  }
  \label{fig:overview}
\end{figure*}
\section{Method}

To identify a set of representative keypoints across a dataset of images, we formulate our approach in an unsupervised framework leveraging conditional diffusion models; see \cref{fig:overview}.
In particular, we utilize the cross-attention maps between the text embeddings and the image features, derived from the latent diffusion model~\cite{stableDiffusion}, and force them to consistently concentrate their activation on highly localized regions within the images.
While \citet{unsupervised_correspondences_using_sd} employed a similar mechanism~(\ie, given a set of keypoint locations in one image, identify correspondences in another image), in this work we seek to identify semantic correspondences across all images within a class-specific dataset~(\eg, human faces), \emph{without any given} knowledge on what and where to focus.
We show that this is possible simply by enforcing locality and equivariance to transformations.

Let us start by quickly reviewing the fundamentals of diffusion models and formalizing the attention maps that we will utilize within these models (\cref{sec:prelim}).
We then detail the objectives used to learn the text embeddings that represent keypoints (\cref{sec:obj}) and discuss important implementation details (\cref{sec:impl}).

\subsection{Attention maps in diffusion networks}
\label{sec:prelim}
Diffusion models are a class of generative models that approximate the data distribution by denoising a base (typically Gaussian) distribution~\cite{ddpm}.
A \textit{latent} diffusion model operates on a latent representation $\latent$ rather than the image itself, with an encoder that maps an image $\image$ into a latent $\latent$, and a decoder that maps  $\latent$ into $\image$.
These models define a \textit{forward} diffusion process, where the latent representation~$\latent$ is gradually transformed into Gaussian noise over a series of $T$ time steps.
The \textit{inverse} process, over a sequence denoising steps $t ={1, \ldots, T}$ predicts the latent noise~$\noise_\params(\latent_t,t)$ which was gradually added in each iteration in order to recover the original (latent) signal.

In our work, we are interested in \textit{conditional} diffusion models, and the explicit attentional relationship between the condition (i.e.~text) and the generated outcome (i.e.~image) that these models learn.
Typically, diffusion models are made conditional on some text~$\prompt$, by providing an embedding $\embedding = \textenc_\params(\prompt)$ from a text encoder~$\textenc_\params$ to the denoiser.
They are then trained to optimize
\begin{align}
\loss{LDM} = \IE_{\latent, t, \noise \sim \calN(0,1)}
\Bigl[
\|\noise - \noise_\params (\latent_t, t, \embedding) \|_2^2
\Bigr],
\label{eq:DM:cond_obj}
\end{align}
where the denoiser $\noise_\params (\latent_t, t, \embedding)$ is typically implemented by a transformer architecture~\cite{ddpm} involving a combination of self-attention and cross-attention layers.
Of our interest here is the cross-attention layers that relate $\embedding$ to $\latent_t$, which we now formalize.

Specifically, in the transformer part of the model, denote $\Phi_\layer^c(\cdot)$ and $\Psi_\layer^c(\cdot)$ as the $c$-th head and the $\layer$-th linear layers of the U-Net.
We calculate the query as~$\query_{\layer}^c = \Phi_\layer^c(\latent_{t{=}1}) \in \IR^{(H \times W) \times D_\layer}$,\footnote{We choose $t{=}1$ where $T{=}50$ steps via hyper-parameter tuning.} and the key from the language embedding~$\key_\layer^c = \Psi_\layer^c(\embedding) \in \IR^{\numTokens \times D_\layer}$, where $\numTokens$ is the number of tokens, $C$ the number of heads in the transformer attention layer, $H$ and $W$ are image height and width at that specific layer in U-Net, and $D_\layer$ the dimensionality of the layer.
Given query and key, the cross-attention map~$\attentionMap_{\layer} \in \IR^{(H \times W) \times \numTokens}$ is then computed via softmax along the $\numTokens$ dimension, and average pooling along the $C$ dimension:
\begin{align}
\attentionMap_\layer(\embedding, \image) &= \IE_c\left[ 
\text{softmax}_\iToken \left({\query_\layer^c \cdot \key_\layer}/{\sqrt{D_\layer}}\right)
\right]
.
\label{eq:attention}
\end{align}
As various layers of the U-Net exhibit distinct levels of semantic understanding, following \citet{unsupervised_correspondences_using_sd}, we collect this information by average pooling across a \textit{selection} of layers:
\begin{align}
\tilde{\attentionMap} =  
\IE_{\layer=7..10} 
\Bigl[
\attentionMap_\layer(\embedding, \image)
\Bigr]
\in \IR^{(H \times W) \times \numTokens}
.
\end{align}
In what follows, to lighten the notation, we drop the attention mask arguments~$(\embedding, \image)$ and write the attention map for the n-th token as $\tilde{\attentionMap}_n$.

\subsection{Optimizing to find the keypoint embeddings}
\label{sec:obj}

To obtain a text embedding that can be used to locate keypoints, for each of them, we simply optimize for two objectives that respectively encourages localization and equivariance to geometric transformations.
We thus write
\begin{equation}
    \loss{total}= \loss{localize} + \lambda_{\text{equiv}}\loss{equiv} 
    ,
    \label{eq:losstotal}
\end{equation}
where we apply $\lambda_{\text{equiv}}{=}10$ to balance the two losses to be in a similar operating range.
Equivariance is enforced in the typical form of learning to be invariant to transformations.
We first quickly detail $\loss{equiv}$, and then discuss how we enforce localization, which is the core of our method.

\paragraph{Equivariance -- $\loss{equiv}$}
To ensure our model's attention mechanism remains consistent across different geometric transformations $\transform$ of the input, we use the typical equivariance loss~\cite{equivariance}:
\begin{equation}
\mathcal{L}_{\text{equiv}} =
\IE_n\ \| \transform^{-1}(\attentionMap_n(\embedding, \transform(\image))) - \attentionMap_n(\embedding, \image) \|^2
\label{eq:equivarianceLoss}
\end{equation}
For $\transform$ we simply utilize minor affine transformations.
We use random rotations between $\pm15$ degrees, translations between $\pm0.25 \times W$, and scaling between 100{--}120\% of the original image size.

\paragraph{Encouraging localization -- $\loss{localize}$}
We encourage localization by forcing $\tilde{\attentionMap}_n$ to be a single-mode Gaussian distribution located at its maximum.
In more details, denoting the Gaussian image that shares the same maximum as $\tilde{\attentionMap}_n$ as $\target_n$, we write
\begin{equation}
\loss{localize} = \IE_n\|\tilde{\attentionMap}_n - \target_n\|^2
.
    \label{eq:loss_local}
\end{equation}
To create the Gaussian images $\target_n$, we first identify the spatial location exhibiting the maximal response within the heatmap corresponding to each token $n$ by taking the argmax:
\begin{equation}
\boldsymbol{\mu}_\iToken = \argmax_{w,h} \:\: \tilde{\attentionMap}_n[h,w]
.
\end{equation}
We then generate a Gaussian image; see \cref{fig:overview}:
\begin{equation}
\target_n = \exp 
\left(
-\frac{\| \mathbf{XY_{coord}} - \boldsymbol{\mu}_\iToken \|_2^2}{2 \sigma^2} 
\right) 
,
\label{eq:gaussian_target}
\end{equation}
where $\mathbf{XY_{coord}}$ is a tensor of image coordinates.

\paragraph{Promoting mutual exclusivity}
It is important to note that while $\loss{localize}$ in \cref{eq:loss_local} at first glance seem to only encourage localization, it also enforces $\tilde{\attentionMap}_n$ to be mutually exclusive for different $n$ because of the softmax operation in \cref{eq:attention}.
Should multiple embeddings become similar, their attention responses in \cref{eq:attention} $\query_\layer^c \cdot \key_\layer$ will become similar, resulting in the softmax of the attention map being a flat response (i.e. deviating from a Gaussian shape).
In other words,~\cref{eq:loss_local} naturally enforces exclusivity with the help of~\cref{eq:attention}.

\paragraph{Stabilizing optimization by working with a subset}
We noticed in our experiments that attention maps $\attentionMap$ for some tokens can be `spread' for some images, \eg, due to occlusions, destabilizing optimization.
We thus opt for a simple solution of looking into the top-K tokens that are local.
Specifically, we apply our losses over $n \in \gaussianSort(\fpsNumSamples)$, which returns the $\fpsNumSamples {\in} \mathbb{N}$ entries with the most spatially localized heatmap responses \footnote{We empirically found that using $\fpsNumSamples{=}25$ works best in general.}, as measured by KL divergence: 
\begin{align}
\gaussianSort(\fpsNumSamples) = \text{ArgTop}_\fpsNumSamples \:\: \{ -\text{KL}(\target_n, \tilde{\attentionMap}_n) \}_{n=1}^{N}
.
\end{align}

\paragraph{Final keypoints}
While $\loss{localize}$ naturally enforces exclusivity, it does not guarantee a complete coverage of the object. 
Thus, after we finish optimizing, we refine the set of keypoints through furthest point sampling using the training images.
Specifically, for each image we write:
\begin{align}
\mathcal{K} = \text{FPS}_\numkeypoints(\{ \boldsymbol{\mu}_i \mid i \in \gaussianSort(\fpsNumSamples) \})
,
\label{eq:topk}
\end{align}
where $\numkeypoints$ is the desired number of keypoints $\numkeypoints < \fpsNumSamples$.
Then, as the set $\mathcal{K}$ differs in each image, we simply choose $\numkeypoints$ tokens that appeared most frequently in $\mathcal{K}$ within the training image set.

\subsection{Implementation details}
\label{sec:impl}

\vspace{-\customparskip}
\paragraph{Test-time ensembling}
At inference time, as in common literature~\cite{unsupervised_correspondences_using_sd,superpoint,openpose},
rather than employing the attention map of the original image, we average the attention maps across multiple augmentations (we use the same transformations for test time augmentation as in \cref{eq:equivarianceLoss}) of the image:
\begin{equation}
\tilde{\attentionMap} \longleftarrow \sum_i \transform_i^{-1} \left( \attentionMap \left( \embedding, \transform_i(\image) \right) \right).
\end{equation}

\paragraph{Upsampling attention maps}
The attention maps in \cref{eq:attention} are typically of low resolution.
Specifically, as we use Stable Diffusion $1.5$~\cite{stableDiffusion}, depending on the layer we extract the attention maps from, they are either $16 \times 16$ or $32 \times 32$.
We thus opt to upsample the query~$\query$ via bicubic interpolation to achieve a standard resolution of $128\times128$.
We have experimented with other upsampling techniques such as the commonly used bilinear sampling or a learned upsampler that is trained alongside, but a simple bicubic upsample was shown to be effective.

\section{Results}
\subsection{Experimental setup}
We evaluate our method on five standard datasets for unsupervised keypoint evaluation:
\begin{itemize}
    \item {\bf CelebA} dataset~\cite{celeba}: 
    A dataset of 202,599 facial images of celebrities. 
    We evaluate both the aligned and non-aligned cases following the standard protocol of omitting images with faces occupying less than 30\% of the image.
    The standard metric for this dataset is to measure the average $\ell_2$ error normalized by the inter-ocular distance.
    \item {\bf CUB-200-2011} dataset~\cite{cub}: 
    This dataset consists of 11,788 bird images.
    We use both the aligned (CUB-aligned) and non-aligned (CUB-all) variants.
    For the non-aligned variants, we further look at CUB-001, CUB-002, and CUB-003, which are specific bird subcategories.
    Notably, these subsets contain only 30 images each---we only use these 30 for training.
    We follow the standard protocol~\cite{lorenz2019unsupervised, choudhury2021unsupervised} and normalize the images to be of $256\times256$.
    The standard metric for this dataset is the mean $\ell_2$ error, normalized by the dimension of the images after normalization.
    \item {\bf Tai-Chi-HD} dataset~\cite{taichi}: 
    This dataset contains 3049 training videos and 285 test videos of people performing Tai-Chi, which shows more diverse poses compared to the other datasets, and is the most challenging among the human pose-centric datasets that we use.
    We follow \citet{siarohin2021motion} and use 500 images for testing and 300 images for training.
    The standard metric for this dataset is to measure the accumulated $\ell_2$ error, with the images standardized to $256\times256$.
    \item {\bf DeepFashion} dataset~\cite{DeepFashion}: 
    This dataset contains 53k images of fashion models, mostly standing with a white background.
    We follow \citet{lorenz2019unsupervised} and only keep full body images. 
    This leaves 10,604 images for training and 1,179 images for testing.
    Also following the baselines, we use keypoints generated by AlphaPose~\cite{alphapose} as ground truth. 
    The standard metric for this dataset is the percentage of correct keypoints (PCK) with a 6-pixel threshold.
    \item {\bf Human~3.6M} dataset~\cite{human36m}: 
    This dataset is of humans performing various actions, comprised of 3.6 million images.
    We follow the standard protocol~\cite{zhang2018unsupervised} and focus on six activities: direction, discussion, posing, waiting, greeting, and walking. 
    We utilize subjects 1, 5, 6, 7, 8, and 9 for training, while subject 11 is reserved for testing. 
    This division yields a training dataset comprising 796,648 images and a testing dataset containing 87,975 images.
    The background for this dataset is also simple, and often masked out with ground-truth masks for evaluation.
    This dataset is also typically heavily pre-processed and aligned when used for unsupervised keypoint evaluation.
    We experiment with the standard pre-processing~\cite{zhang2018unsupervised, lorenz2019unsupervised} and also a relaxed version of our own.
    To relax the alignment, we crop a square bounding box such that the margin from the bounding box to the person is $100$ pixels, which on average corresponds to the person's height being $2/3$ of the crop.
    We further add a uniform random translation up to $100$ pixels (same as the margin) to remove the central bias.
    Example crops are visualized in \cref{fig:unaligned_human36m_keypoints}.
    The standard metric for this dataset is the $\ell_2$ error after normalizing the image resolution to 128$\times$128.
\end{itemize}
\noindent
Note that each dataset comes with its own metric. 
To make results more comparable across the human pose datasets, we report both their original metrics as well as the $\ell_2$ error when normalizing the image resolution to 128$\times$128.

\paragraph{Regressing human-annotated landmarks}
To evaluate the quality of unsupervised keypoints, one must relate them with human-annotated landmarks.
As in prior research~\cite{thewlis2017unsupervised}, we use linear regression (without bias) to relate between unsupervised keypoints and human-annotated landmarks.

\paragraph{Number of keypoints and hyperparameters}
For each method, we use the standard number of unsupervised keypoints defined for each evaluation protocol---we denote them in our Tables.
We use the same hyperparameter for all our experiments as introduced in \cref{sec:obj}, except for the number of optimization iterations.
We optimize the embeddings for 10k iterations, except for the human pose datasets, for which we optimize 500 iterations.
To find the number of optimization rounds we use a 10\% validation subset from the training data.
While we observed our results on the validation subset to improve consistently for most datasets, we found 10k to give a reasonable optimization time of two hours on an RTX 3090.
For the human pose dataset, we found optimization to have converged already at 500 iterations on our validation split.

\subsection{Experimental results}

\begin{table}
    \centering
    \setlength\tabcolsep{6pt} %
    \resizebox{\linewidth}{!}{
    \begin{tabular}{@{}l rrr@{}}
        \toprule
        Method                       & \makecell[l]{Aligned\\($\numkeypoints$=10)$\downarrow$}  & \makecell[l]{Wild\\ ($\numkeypoints$=4) $\downarrow$} & \makecell[l]{Wild\\ ($\numkeypoints$=8) $\downarrow$} \\
        \midrule
        Thewlis et al. \cite{thewlis2017unsupervised}         &  7.95                      & -                       & 31.30                 \\
        Zhang et al. \cite{zhang2018unsupervised}           &  3.46                      & -                       & 40.82                 \\
        LatentKeypointGAN \cite{latentkeypointgan}       &  5.85                      & 25.81                 & 21.90                 \\
        Lorenz et al. \cite{lorenz2019unsupervised}           &  3.24                      & 15.49                & 11.41                \\
        IMM \cite{jakab2018unsupervised}                     &  \textbf{3.19}             & 19.42                & 8.74                 \\
        LatentKeypointGAN-tuned \cite{latentkeypointgan} &  3.31                      & 12.10                 & 5.63                  \\
        Autolink  \cite{autolink}           &  3.92                 & 7.72             & 5.66             \\
        Autolink $\dagger$ \cite{autolink}   &  3.54                      & 6.11        & 5.24                  \\
        \textbf{Our method}                         &  3.60                      & \textbf{5.24}                 & \textbf{4.35}        \\
        \bottomrule
    \end{tabular}
    }
    \caption{
    \textbf{Quantitave results for the CelebA dataset --}
    we report results with the standard metrics. 
    Our method performs best for non-aligned cases and is comparable to the state of the art for the aligned case.
    $\dagger$ symbol represents the thickness-tuned variant.
    }
    \label{tab:celeba}
\end{table}
\begin{table}
    \centering
    \setlength\tabcolsep{2pt} %
    \resizebox{\linewidth}{!}{
    \begin{tabular}{@{}l l rrrrr@{}}
        \toprule
        Method                     & Supervision     
        & \makecell{CUB-aligned\\($\numkeypoints$=10) $\downarrow$}
        & \makecell{CUB-001\\($\numkeypoints$=4) $\downarrow$} 
        & \makecell{CUB-002\\($\numkeypoints$=4) $\downarrow$} 
        & \makecell{CUB-003\\($\numkeypoints$=4) $\downarrow$} 
        & \makecell{CUB-all\\($\numkeypoints$=4) $\downarrow$} \\
        \midrule
        SCOPS \cite{scoops}                 & GT silhouette   & {-}                        & 18.3                   & 17.7                   & 17.0                   & 12.6                   \\
        Choudhury et al. \cite{choudhury2021unsupervised}      & GT silhouette   & {-}                        & 11.3                   & 15.0                   & 10.6                   & 9.2                    \\
        DFF \cite{collins2018deep}                   & testing dataset & {-}                        & 22.4                   & 21.6                   & 22.0                   & {-}                    \\
        SCOPS \cite{scoops}                 & saliency maps   & {-}                        & 18.5                   & 18.8                   & 21.1                   & {-}                    \\
        \midrule
        Lorenz et al. \cite{lorenz2019unsupervised}         & unsupervised    & 3.91                       & {-}                    & {-}                    & {-}                    & {-}                    \\
        ULD \cite{zhang2018unsupervised, thewlis2017unsupervised}             & unsupervised    & {-}                        & 30.1                   & 29.4                   & 28.2                   & {-}                    \\
        Zhang et al. \cite{zhang2018unsupervised}         & unsupervised    & 5.36                       & 26.9                   & 27.6                   & 27.1                   & 22.4                   \\
        LatentKeypointGAN \cite{latentkeypointgan}     & unsupervised    & 5.21                       & 22.6                   & 29.1                   & 21.2                   & 14.7                   \\
        GANSeg \cite{ganseg}                & unsupervised    & \textbf{3.23}                       & 22.1                   & 22.3                   & 21.5                   & 12.1                   \\
        Autolink \cite{autolink}        & unsupervised    & 4.15                       & 20.6                   & 20.3                   & 19.7                   & 11.6                   \\
        Autolink $\dagger$ \cite{autolink} & unsupervised    & 3.51                       & 20.2                   & 19.2                   & 18.5                   & 11.3                   \\
\textbf{Our method}                      & unsupervised    & 5.06                      & \textbf{10.5}          & \textbf{11.1}         & \textbf{10.3}          & \textbf{5.4}          \\
        \bottomrule
    \end{tabular}
    }
    \caption{
    \textbf{Quantitave results for the CUB-200-2011 dataset --}
    we report results with the standard metrics. 
    Except for the CUB-aligned case, our method performs nearly twice better than the compared methods, even outperforming \citet{choudhury2021unsupervised}, which is supervised with ground-truth silhouettes.
    $\dagger$ represents the thickness-tuned variant.
    }
    \label{tab:cub}
\end{table}
\begin{table}
    \centering
    \setlength\tabcolsep{2pt} %
    \resizebox{\linewidth}{!}{
    \begin{tabular}{@{}ll ccc@{}}
        \toprule
        Method & Supervision & \makecell{Human 3.6M \\ ($\numkeypoints$=16) \\ $\ell_2$ standard / unaligned $\downarrow$} & \makecell{DeepFashion \\ ($\numkeypoints$=16)\\ PCK$\uparrow$ / Rel. $\ell_2\downarrow$} & \makecell{Tai-Chi-HD \\ ($\numkeypoints$=10)\\ Cum $\ell_2\downarrow$ / Rel. $\ell_2\downarrow$} \\
        \midrule
        Newell et al. \cite{jakab2020self} & paired gt & 2.16 / {-} & - & - \\
        DFF \cite{collins2018deep} & testing dataset & - & - & 494.48 / 14.78 \\
        SCOPS \cite{scoops} & saliency maps & - & - &  411.38 / 12.29 \\
        \hline
        Jakab et al. \cite{jakab2020self} & video* & 2.73 / {-} & - & - \\
        Siarohin et al. \cite{siarohin2021motion} & videos & - & - &  389.78 / 11.65 \\
        Zhang et al. \cite{zhang2022self} & videos & - & - &  343.67 / 10.27 \\
        Zhang et al. \cite{zhang2018unsupervised} & videos & 4.14 / {-} & - & - \\
        Schmidtke et al. \cite{schmidtke2021unsupervised} & video* & 3.31 / {-} & - & - \\
        Sun et al. \cite{sun2022self} & videos & 2.53 / {-} & - & - \\
        \hline
        Thewlis et al. \cite{thewlis2017unsupervised} & unsupervised & 7.51 / {-} & - & - \\
        Zhang et al. \cite{zhang2018unsupervised} & unsupervised & 4.91 / {-} & - & - \\
        LatentKeypointGAN \cite{latentkeypointgan} & unsupervised & - & 49\% &  437.69 / 13.08 \\
        Lorenz et al. \cite{lorenz2019unsupervised} & unsupervised & 2.79 / {-} & 57\% & - \\
        GANSeg \cite{ganseg} & unsupervised & - & 59\% &  417.17 / 12.47 \\
        autolink \cite{autolink} & unsupervised & 2.81 / 7.59 & 65\% &  337.50 / 10.08 \\
        autolink $\dagger$ \cite{autolink} & unsupervised & \textbf{2.76} / {-} & 66\% &  316.10 / 9.45 \\
        \textbf{Our method} & unsupervised & 4.45 / \textbf{5.77} & \textbf{70\%}/6.46 & \textbf{234.89 / 7.02} \\
        \bottomrule
    \end{tabular}
    }
    \caption{
    {\bf Quantitative results for human pose datasets --}
    We report results for the Human 3.6M dataset, Deep Fashion dataset, and the Challenging Tai-Chi-HD datasets. We report both standard metrics for each dataset and the relative $\ell_2$ error after normalizing images to $128\times128$. Our method, except for the Human 3.6M dataset that is heavily pre-processed, outperforms all baselines. This includes, for the challenging Tai-Chi-HD dataset, supervised ones.
    * denotes additional supervision (\citet{jakab2020self} uses unpaired ground truth, and Schmidtke~\etal~\cite{schmidtke2021unsupervised} use the T-pose).
    The $\dagger$ symbol represents the thickness-tuned variant for Autolink.
    }
    \label{tab:humans}
\end{table}

\def \figfivew {0.49} %
\begin{figure*}[t]
  \centering
  \begin{subfigure}[b]{\figfivew\textwidth}
    \centering
    \includegraphics[width=\linewidth]{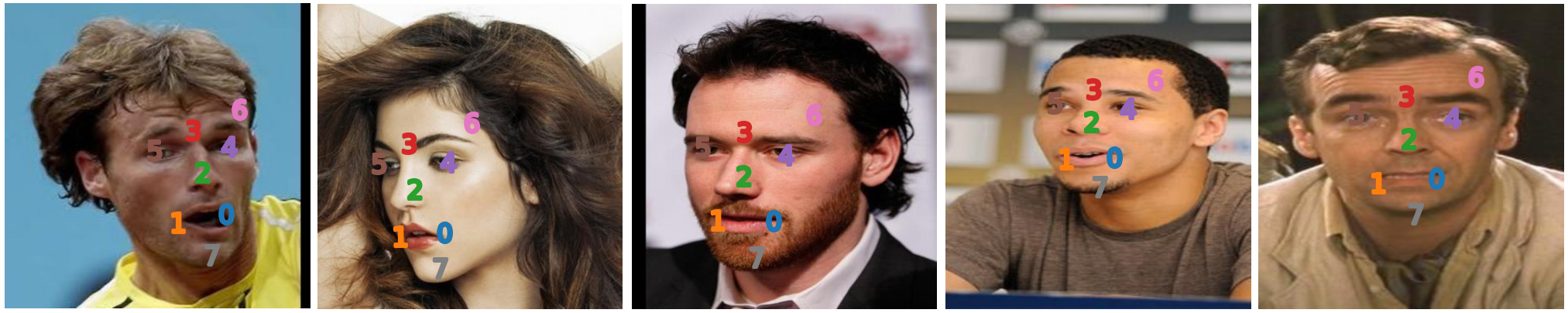}
    \caption{CelebA dataset keypoints}
    \label{fig:celeba_keypoints}
  \end{subfigure}
  \hfill %
  \begin{subfigure}[b]{\figfivew\textwidth}
    \centering
    \includegraphics[width=\linewidth]{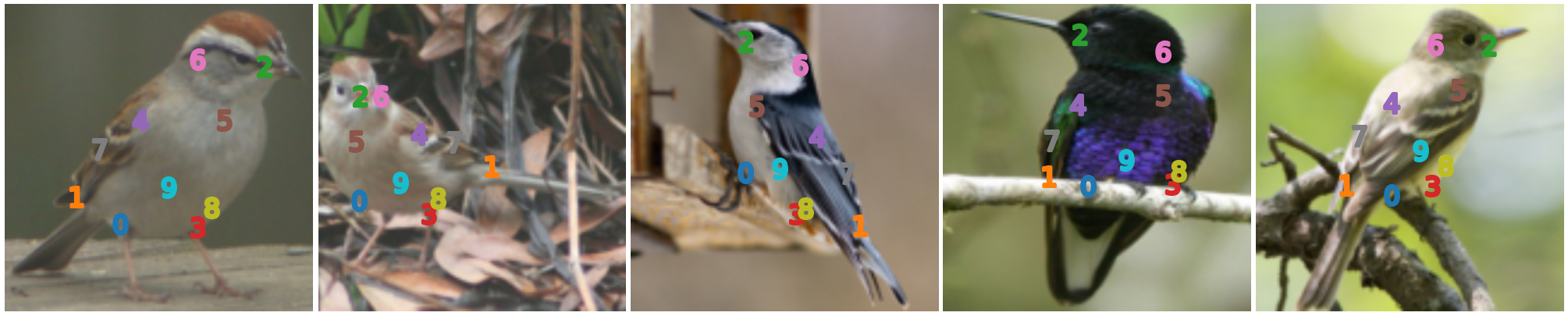}
    \caption{CUB-200-2011 dataset keypoints}
    \label{fig:cub_keypoints}
  \end{subfigure}

  \begin{subfigure}[b]{\figfivew\textwidth}
    \centering
    \includegraphics[width=\linewidth]{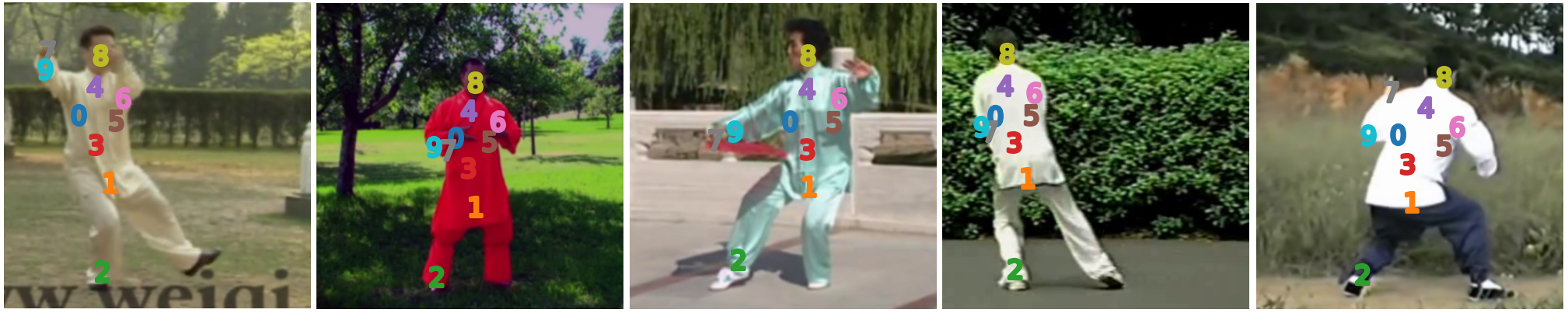}
    \caption{Tai-Chi-HD dataset keypoints}
    \label{fig:taichi_keypoints}
  \end{subfigure}
  \hfill
  \begin{subfigure}[b]{\figfivew\textwidth}
    \centering
    \includegraphics[width=\linewidth]{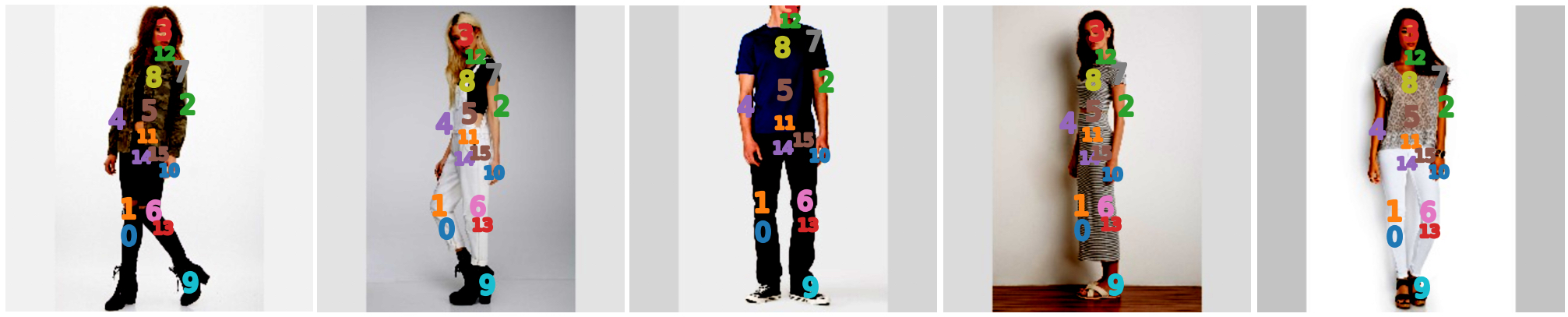}
    \caption{DeepFashion dataset keypoints}
    \label{fig:deepfashion_keypoints}
  \end{subfigure}

  \begin{subfigure}[b]{\figfivew\textwidth}
    \centering
    \includegraphics[width=\linewidth]{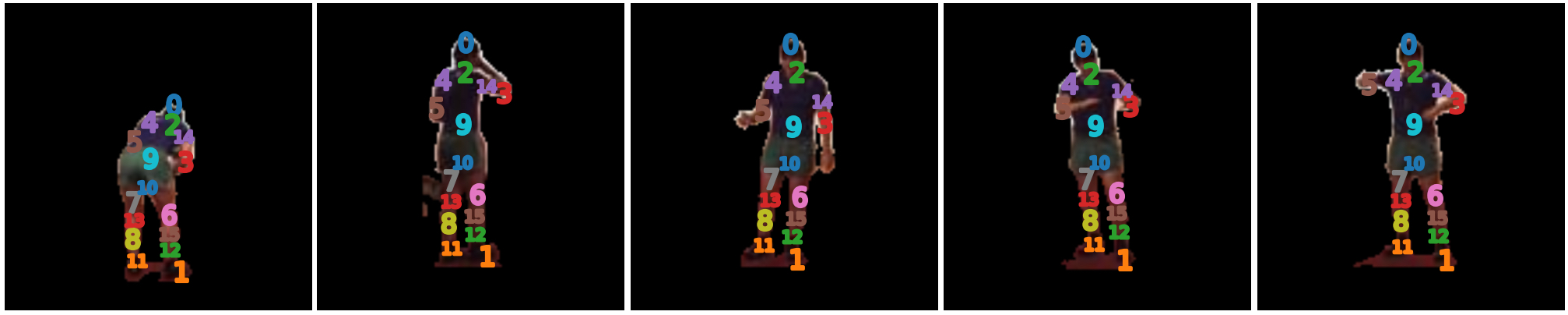}
    \caption{Human 3.6M dataset keypoints}
    \label{fig:human36m_keypoints}
  \end{subfigure}
  \hfill
  \begin{subfigure}[b]{\figfivew\textwidth}
    \centering
    \includegraphics[width=\linewidth]{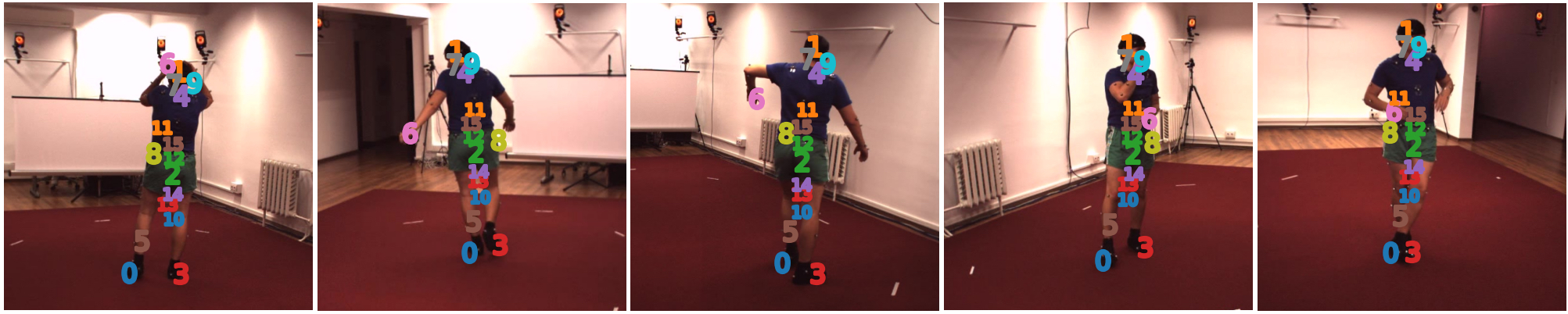}
    \caption{Unaligned Human 3.6M dataset keypoints}
    \label{fig:unaligned_human36m_keypoints}
  \end{subfigure}

  \caption{
    {\bf Qualitative examples of unsupervised keypoints --}
    we show our learned keypoints for the CelebA, CUB-200-2011, Tai-Chi-HD, DeepFashion, and Human 3.6M datasets (both for cropped and masked as well as our relaxed version).
    Note how our keypoints are consistent despite the variability.
    Our method significantly outperforms other baselines, especially for the challenging Tai-Chi-HD dataset and the CUB subsets.
  }
  \label{fig:unsupervised_keypoints}
\end{figure*}

\vspace{-\customparskip}
\paragraph{Quantiative results -- \cref{tab:celeba,tab:cub,tab:humans}}
We report our results for each dataset in \cref{tab:celeba,tab:cub,tab:humans}.
As shown, except for the case when data is heavily processed and aligned (CelebA aligned in \cref{tab:celeba}, CUB-aligned in \cref{tab:cub}, and Human 3.6M in \cref{tab:humans}), our method significantly outperforms the state of the art. 
The most visible gains are for the Tai-Chi-HD dataset, the most challenging among human pose datasets, and on CUB unaligned datasets.
For the CUB dataset and the Tai-Chi-HD datasets, we outperform even those that have been supervised with silhouettes or saliency maps.

We note that our primary focus is on unaligned cases, as we argue that they represent more how keypoints would be used in real-world applications---most real-world datasets are unaligned except for specific classes of objects.
Moreover, methods focusing on aligned settings use strong locational priors, and as shown by their results in the unaligned setup---CelebA in the wild, non-aligned cases of Human 3.6M and CUB-200-2011, and Tai-Chi-HD---may perform significantly worse once this alignment prior is broken.
Given that the performance of our method, even in the aligned case, is not too far off from methods that utilize alignment, we suspect a more in-depth tuning of our method may make our method outperform these methods, but we leave this as future work.

Finally, also note that for CUB-001, CUB-002, and CUB-003, these datasets are small.
These datasets are non-aligned, have a large variability between individual images, and only contain 30 images each in the training set.
Our method, \emph{just from 30 images}, successfully identifies keypoints.
These results highlight the potential of leveraging emergent (prior) knowledge within Stable Diffusion~\cite{stableDiffusion}.

\paragraph{Qualitative results -- \cref{fig:unsupervised_keypoints}}
We provide example visualizations of our unsupervised keypoints in \cref{fig:unsupervised_keypoints}.
As shown, our method discovers keypoints that are consistently localized across the dataset, despite the wide appearance variety.

\subsection{Ablation study}
\label{sec:ablation}

\begin{table}
    \centering
    \setlength\tabcolsep{18pt} %
    \resizebox{\linewidth}{!}{
    \begin{tabular}{@{}l r@{}}
        \toprule
        Variant & Normalized $\ell_2$ \\
        \midrule
        \textbf{Full (Our method)}  & 5.4 \\
        Without test time ensembling & 5.6 \\
        Without furthest point sampling   & 6.4 \\
        Without upsampling the query $\query$ & 8.0 \\
        Without equivariance  & 22.2 \\
        \bottomrule
    \end{tabular}
    }
    \caption{
    {\bf Ablation results -- }
    we report the effect of each of our design choices can be seen on the CUB-all dataset.
    All components contribute to the final performance.
    }
    \label{tab:ablation}
\end{table}

We perform an ablation study for various design choices of our method on the CUB-all dataset.
We report the performance of our method with different components disabled in \cref{tab:ablation}.
As shown, all components contribute to the final performance.
Test-time ensembling enhances performance, but the computation cost linearly scales.
We choose, ten augmentations, which provide a good compromise between computation time and accuracy.
To remove furthest point sampling we set $\fpsNumSamples{=}\numkeypoints$, which then makes furthest point sampling select all samples.
While this causes points to be more grouped, it still provides reasonable performance.
To remove upsampling we instead upsample $\attentionMap$ to the size of the target image, effectively having the attention map build at lower resolutions, sometimes as low as $16{\times}16$.
This results in significant degradation in performance.
$\loss{equiv}$ is essential, as without it, the tokens can `cheat' and simply opt to learn fixed positions on the image.

\paragraph{Number of training images.}
Inspired by our results for the small subsets of CUB-200-2011 dataset, we investigate the impact that the number of images that we use to find keypoints has on our results.
We thus optimized our keypoints only with 100 images for CelebA non-aligned setup.
Surprisingly, we achieve 5.33 $K{=}8$, which is comparable to the state of the art.
This demonstrates once more how our method is able to leverage information that is already learned in Stable Diffusion~\cite{stableDiffusion} to find keypoints.

\begin{figure*}
  \centering
  \begin{subfigure}{0.495\linewidth}
    \centering
    \includegraphics[width=\linewidth]{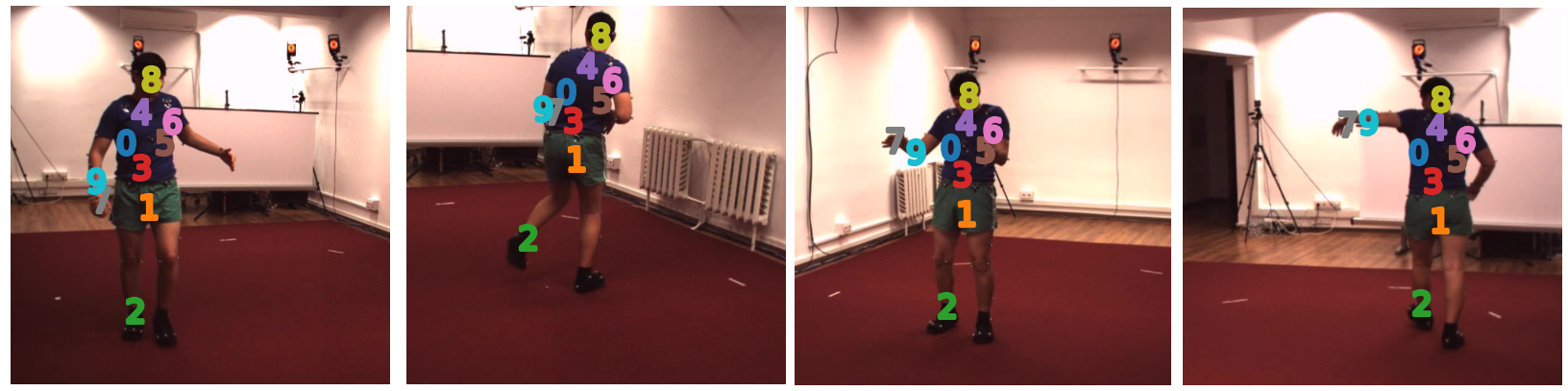}
    \caption{Applying Tai-Chi-HD tokens to Human 3.6M}
    \label{fig:taichi_on_human36m}
  \end{subfigure}
  \begin{subfigure}{0.495\linewidth}
    \centering
    \includegraphics[width=\linewidth]{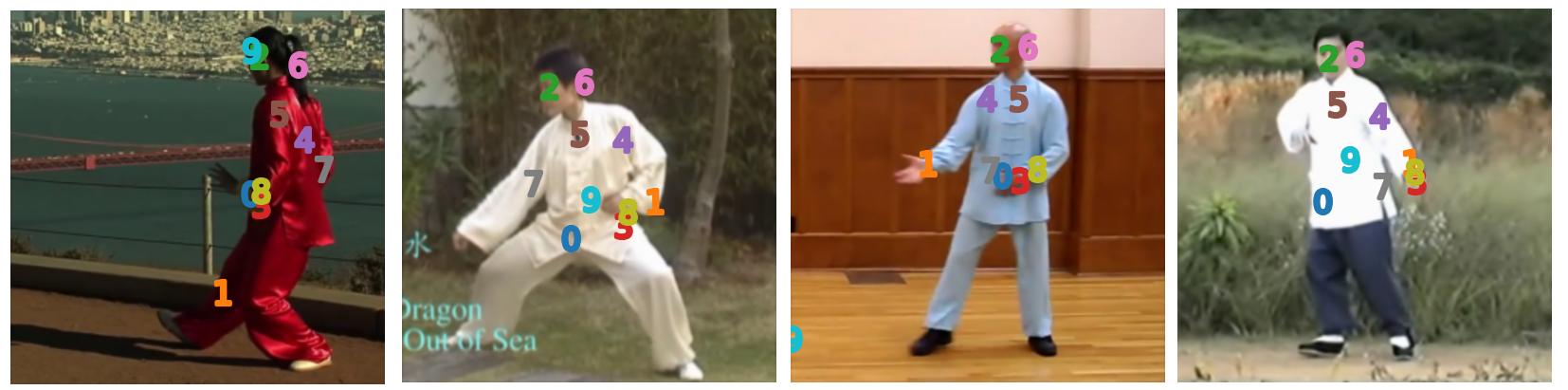}
    \caption{Applying CUB-200-2011 tokens to Tai-Chi-HD}
    \label{fig:cub_on_taichi}
  \end{subfigure}

  \begin{subfigure}{0.495\linewidth}
    \centering
    \includegraphics[width=\linewidth]{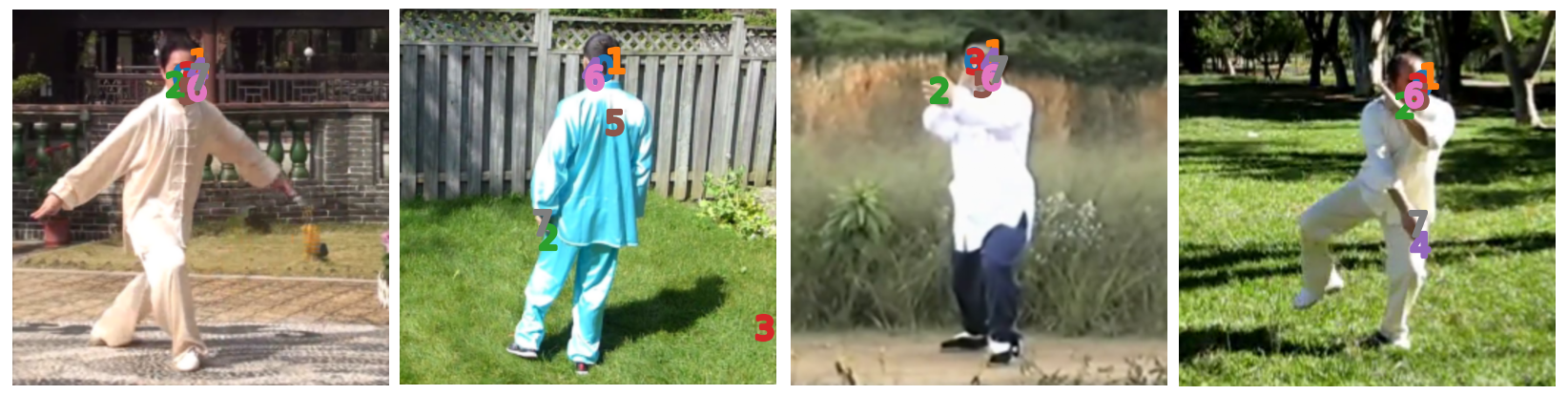}
    \caption{Applying CelebA tokens to Tai-Chi-HD }
    \label{fig:celeba_on_taichi}
  \end{subfigure}
  \begin{subfigure}{0.495\linewidth}
    \centering
    \includegraphics[width=\linewidth]{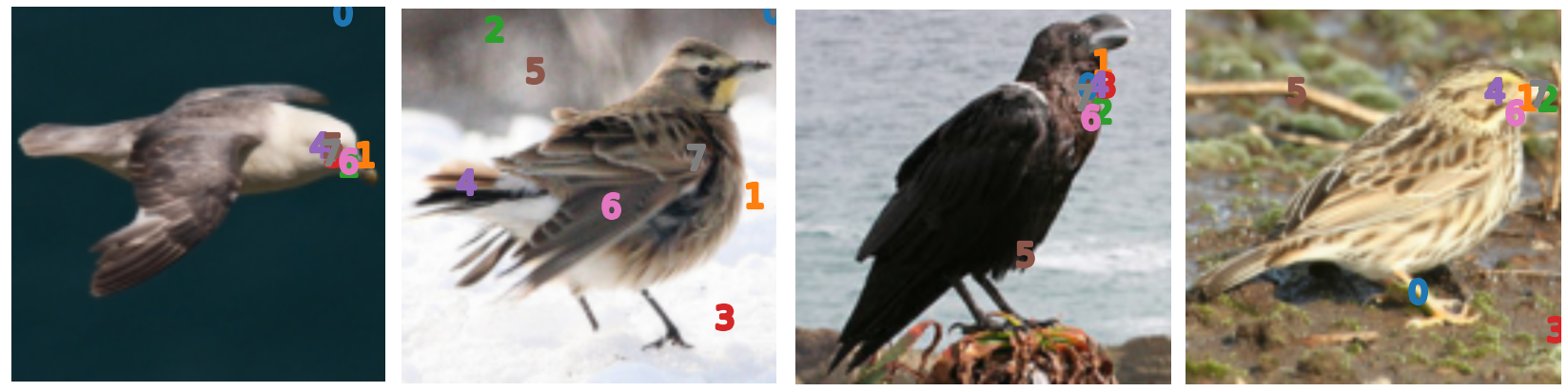}
    \caption{Applying CelebA tokens to CUB-200-2011}
    \label{fig:celeba_on_cub}
  \end{subfigure}

  \caption{
  {\bf Generalization --}
  we apply our learned text tokens (keypoints) to images from other datasets, including those that are of completely different domains.
  Our tokens generalize well for data of similar type, and surprisingly well even for some extreme cases.
  }
  \label{fig:applyingOOD}
\end{figure*}

\subsection{Generalization}
We further test the generalization capacity of our learned keypoints.
As they are effectively text embeddings, we can simply apply them to any image, including those completely outside of the training domain.
We quantitatively evaluate our method and the previous best-performing method Autolink \cite{autolink}.
We find that even in these generalization experiments, our keypoints reach performance comparable to data-specific keypoints.
Applying Tai-Chi-HD tokens to unaligned Human 3.6M achieves state-of-the-art performance despite using fewer keypoints (K=10 vs K=16). 
We also perform on par with the previous state of the art when we apply CUB-200-2011 tokens to Tai-Chi-HD---a case where the dataset gap is not only about the appearance but also beyond classes. 
Despite the drastic gap, our keypoints perform extremely well, leveraging the generalization power of large pre-trained diffusion models. 

We show qualitative examples in \cref{fig:applyingOOD}.
As shown, even when applied to different datasets, they look reasonable.
For example, in \cref{fig:taichi_on_human36m}, when applying Tai-Chi-HD tokens to Human 3.6M, the tokens respond to the same locations on the human body as in Tai-Chi-HD.
Surprising was when we applied CUB-200-2011 tokens to Tai-Chi-HD in \cref{fig:cub_on_taichi}---they still responded to the body of the human being, reasonably consistently, although these tokens were trained to respond to birds.
Of note are tokens two and six, which correspond to the front and back of the bird heads in \cref{fig:unsupervised_keypoints}---they also reply to the front and back of human heads.
Applying CelebA tokens to Tai-Chi-HD in \cref{fig:celeba_on_taichi} also shows interesting outcomes, as tokens generally respond to human faces, despite the scale being drastically different between the two datasets.
Finally, applying CelebA tokens to the CUB-200-2011 dataset in \cref{fig:celeba_on_cub} shows mixed results---when it is `successful' it focuses also on the faces of the bird, when it fails, it fails completely.
These results hint that the keypoints (tokens) we have learned carry semantic meanings, as expected.
We note that none of the baselines that we compare against are able to generalize beyond the dataset they were trained for.

\begin{centering}
\begin{table} %
\centering
\setlength\tabcolsep{6pt} %
\resizebox{\linewidth}{!}{
\begin{tabular}{ccccc}
\hline
 & \makecell[l]{Tai-Chi-HD\\$\rightarrow$unaligned\\Human3.6m\\($\numkeypoints$=10)$\downarrow$} & \makecell[l]{CUB-200-2011\\$\rightarrow$Tai-Chi-HD\\($\numkeypoints$=10)$\downarrow$\\Cum $\ell_2\downarrow$ / Rel. $\ell_2\downarrow$} & \makecell[l]{CelebA\\$\rightarrow$Tai-Chi-HD\\($\numkeypoints$=8)$\downarrow$\\Cum $\ell_2\downarrow$ / Rel. $\ell_2\downarrow$} & \makecell[l]{CelebA\\$\rightarrow$CUB-200-2011\\($\numkeypoints$=8)$\downarrow$} \\
\hline
Ours & \textbf{4.88} & \textbf{317.94 / 9.50} & - / \textbf{8.6} & \textbf{18.60} \\
Autolink \cite{autolink} & 16.92 & 535.61 / 16.00 & - / 28.2 & 22.56 \\
\hline
\end{tabular}
}
\caption{
{\bf Generalization -- } 
we quantitatively evaluate the performance of our keypoints on other datasets.
Our Tai-Chi-HD keypoints applied to the unaligned Human3.6m setting reach state-of-the-art performance. 
Interestingly, our CUB keypoints applied to Tai-Chi-HD are on par with the previous state of the art, despite the differences between these datasets.
}
\end{table}
\end{centering}

\section{Conclusions}

We have proposed a novel method to find unsupervised keypoints using pre-trained text-to-image diffusion models.
Given a set of images of a certain object, we propose to optimize the text embeddings (tokens) such that the cross-attention maps within diffusion models become localized as Gaussians with a small standard deviation.
By doing so, we find text tokens that can be used to extract keypoints by extracting the maxima of the attention maps.
We have shown that our method, on multiple datasets, under the challenging un-aligned setup, significantly outperforms the state of the art.
We have further demonstrated that these tokens are also generalizable.
\section{Acknowledgments}

The authors would like to thank Cristina Vasconcelos for her constructive feedback during the preparation of this manuscript. Additionally, we extend our gratitude to David Fleet for his approval and support of this work.

This work was supported in part by the Natural Sciences and Engineering Research Council of Canada (NSERC) Discovery Grant, NSERC Collaborative Research and Development Grant, Google, Digital Research Alliance of Canada, and Advanced Research Computing at the University of British Columbia.

{
    \small
    \bibliographystyle{ieeenat_fullname}
    \bibliography{macros.bib, bibliography}

\begin{thebibliography}{69}
\providecommand{\natexlab}[1]{#1}
\providecommand{\url}[1]{\texttt{#1}}
\expandafter\ifx\csname urlstyle\endcsname\relax
  \providecommand{\doi}[1]{doi: #1}\else
  \providecommand{\doi}{doi: \begingroup \urlstyle{rm}\Url}\fi

\bibitem[Azizi et~al.(2023)Azizi, Kornblith, Saharia, Norouzi, and Fleet]{azizi2023synthetic}
Shekoofeh Azizi, Simon Kornblith, Chitwan Saharia, Mohammad Norouzi, and David~J Fleet.
\newblock Synthetic data from diffusion models improves imagenet classification.
\newblock \emph{Transactions on Machine Learning Research}, 2023.

\bibitem[Baranchuk et~al.(2021)Baranchuk, Rubachev, Voynov, Khrulkov, and Babenko]{baranchuk2021label}
Dmitry Baranchuk, Ivan Rubachev, Andrey Voynov, Valentin Khrulkov, and Artem Babenko.
\newblock Label-efficient semantic segmentation with diffusion models.
\newblock \emph{International Conference on Learning Representations}, 2021.

\bibitem[Cao et~al.(2017)Cao, Simon, Wei, and Sheikh]{openpose}
Zhe Cao, Tomas Simon, Shih-En Wei, and Yaser Sheikh.
\newblock Realtime multi-person 2d pose estimation using part affinity fields.
\newblock In \emph{Proceedings of the IEEE/CVF Conference on Computer Vision and Pattern Recognition}, 2017.

\bibitem[Chen et~al.(2022)Chen, Sun, Song, and Luo]{chen2022diffusiondet}
Shoufa Chen, Peize Sun, Yibing Song, and Ping Luo.
\newblock Diffusiondet: Diffusion model for object detection.
\newblock \emph{Proceedings of the IEEE International Conference on Computer Vision}, 2022.

\bibitem[Chen et~al.(2020)Chen, Yu, Tu, Lyu, Tang, Ou, Fu, and Xue]{chen2020survey}
Weiya Chen, Chenchen Yu, Chenyu Tu, Zehua Lyu, Jing Tang, Shiqi Ou, Yan Fu, and Zhidong Xue.
\newblock A survey on hand pose estimation with wearable sensors and computer-vision-based methods.
\newblock \emph{Sensors}, 2020.

\bibitem[Choudhury et~al.(2021)Choudhury, Laina, Rupprecht, and Vedaldi]{choudhury2021unsupervised}
Subhabrata Choudhury, Iro Laina, Christian Rupprecht, and Andrea Vedaldi.
\newblock Unsupervised part discovery from contrastive reconstruction.
\newblock \emph{Advances in Neural Information Processing Systems}, 2021.

\bibitem[Chowdhery et~al.(2022)Chowdhery, Narang, Devlin, Bosma, Mishra, Roberts, Barham, Chung, Sutton, Gehrmann, et~al.]{palm}
Aakanksha Chowdhery, Sharan Narang, Jacob Devlin, Maarten Bosma, Gaurav Mishra, Adam Roberts, Paul Barham, Hyung~Won Chung, Charles Sutton, Sebastian Gehrmann, et~al.
\newblock Palm: Scaling language modeling with pathways.
\newblock \emph{arXiv Preprint}, 2022.

\bibitem[Clark and Jaini(2023)]{clark2023text}
Kevin Clark and Priyank Jaini.
\newblock Text-to-image diffusion models are zero-shot classifiers.
\newblock \emph{International Conference on Learning Representations}, 2023.

\bibitem[Collins et~al.(2018)Collins, Achanta, and Susstrunk]{collins2018deep}
Edo Collins, Radhakrishna Achanta, and Sabine Susstrunk.
\newblock Deep feature factorization for concept discovery.
\newblock In \emph{Proceedings of the European Conference on Computer Vision}, 2018.

\bibitem[DeTone et~al.(2018)DeTone, Malisiewicz, and Rabinovich]{superpoint}
Daniel DeTone, Tomasz Malisiewicz, and Andrew Rabinovich.
\newblock Superpoint: Self-supervised interest point detection and description.
\newblock In \emph{Proceedings of the IEEE Conference on Computer Vision and Pattern Recognition Workshops}, 2018.

\bibitem[Fang et~al.(2022)Fang, Li, Tang, Xu, Zhu, Xiu, Li, and Lu]{alphapose}
Hao-Shu Fang, Jiefeng Li, Hongyang Tang, Chao Xu, Haoyi Zhu, Yuliang Xiu, Yong-Lu Li, and Cewu Lu.
\newblock Alphapose: Whole-body regional multi-person pose estimation and tracking in real-time, 2022.

\bibitem[He et~al.(2021)He, Wandt, and Rhodin]{latentkeypointgan}
Xingzhe He, Bastian Wandt, and Helge Rhodin.
\newblock Latentkeypointgan: Controlling gans via latent keypoints.
\newblock \emph{International Conference on Learning Representations}, 2021.

\bibitem[He et~al.(2022{\natexlab{a}})He, Wandt, and Rhodin]{autolink}
Xingzhe He, Bastian Wandt, and Helge Rhodin.
\newblock Autolink: Self-supervised learning of human skeletons and object outlines by linking keypoints.
\newblock In \emph{Advances in Neural Information Processing Systems}, 2022{\natexlab{a}}.

\bibitem[He et~al.(2022{\natexlab{b}})He, Wandt, and Rhodin]{ganseg}
Xingzhe He, Bastian Wandt, and Helge Rhodin.
\newblock Ganseg: Learning to segment by unsupervised hierarchical image generation.
\newblock In \emph{Proceedings of the IEEE/CVF Conference on Computer Vision and Pattern Recognition}, 2022{\natexlab{b}}.

\bibitem[Hedlin et~al.(2023)Hedlin, Sharma, Mahajan, Isack, Kar, Tagliasacchi, and Yi]{unsupervised_correspondences_using_sd}
Eric Hedlin, Gopal Sharma, Shweta Mahajan, Hossam Isack, Abhishek Kar, Andrea Tagliasacchi, and Kwang~Moo Yi.
\newblock Unsupervised semantic correspondence using stable diffusion.
\newblock \emph{Advances in Neural Information Processing Systems}, 2023.

\bibitem[Hung et~al.(2019)Hung, Jampani, Liu, Molchanov, Yang, and Kautz]{scoops}
Wei-Chih Hung, Varun Jampani, Sifei Liu, Pavlo Molchanov, Ming-Hsuan Yang, and Jan Kautz.
\newblock Scops: Self-supervised co-part segmentation.
\newblock In \emph{Conference on Computer Vision and Pattern Recognition}, 2019.

\bibitem[Ionescu et~al.(2013)Ionescu, Papava, Olaru, and Sminchisescu]{human36m}
Catalin Ionescu, Dragos Papava, Vlad Olaru, and Cristian Sminchisescu.
\newblock Human3. 6m: Large scale datasets and predictive methods for 3d human sensing in natural environments.
\newblock \emph{IEEE Transactions on Pattern Analysis and Machine Intelligence}, 2013.

\bibitem[Jabberi et~al.(2023)Jabberi, Wali, Chaudhuri, and Alimi]{jabberi202368}
Marwa Jabberi, Ali Wali, Bidyut~Baran Chaudhuri, and Adel~M Alimi.
\newblock 68 landmarks are efficient for 3d face alignment: what about more? 3d face alignment method applied to face recognition.
\newblock \emph{Multimedia Tools and Applications}, 2023.

\bibitem[Jakab et~al.(2018)Jakab, Gupta, Bilen, and Vedaldi]{jakab2018unsupervised}
Tomas Jakab, Ankush Gupta, Hakan Bilen, and Andrea Vedaldi.
\newblock Unsupervised learning of object landmarks through conditional image generation.
\newblock \emph{Advances in Neural Information Processing Systems}, 2018.

\bibitem[Jakab et~al.(2020)Jakab, Gupta, Bilen, and Vedaldi]{jakab2020self}
Tomas Jakab, Ankush Gupta, Hakan Bilen, and Andrea Vedaldi.
\newblock Self-supervised learning of interpretable keypoints from unlabelled videos.
\newblock In \emph{Proceedings of the IEEE/CVF Conference on Computer Vision and Pattern Recognition}, 2020.

\bibitem[Jau et~al.(2020)Jau, Zhu, Su, and Chandraker]{jau2020deep}
You-Yi Jau, Rui Zhu, Hao Su, and Manmohan Chandraker.
\newblock Deep keypoint-based camera pose estimation with geometric constraints.
\newblock In \emph{International Conference on Intelligent Robots and Systems}, 2020.

\bibitem[Jiang et~al.(2022)Jiang, Lee, Teotia, and Ostadabbas]{jiang2022animal}
Le Jiang, Caleb Lee, Divyang Teotia, and Sarah Ostadabbas.
\newblock Animal pose estimation: A closer look at the state-of-the-art, existing gaps and opportunities.
\newblock \emph{Computer Vision and Image Understanding}, 2022.

\bibitem[Jin et~al.(2022)Jin, Sun, Hosang, Trulls, and Yi]{tusk}
Yuhe Jin, Weiwei Sun, Jan Hosang, Eduard Trulls, and Kwang~Moo Yi.
\newblock Tusk: Task-agnostic unsupervised keypoints.
\newblock \emph{Advances in Neural Information Processing Systems}, 2022.

\bibitem[Khani et~al.(2023)Khani, Taghanaki, Sanghi, Amiri, and Hamarneh]{slime}
Aliasghar Khani, Saeid~Asgari Taghanaki, Aditya Sanghi, Ali~Mahdavi Amiri, and Ghassan Hamarneh.
\newblock Slime: Segment like me.
\newblock \emph{arXiv Preprint}, 2023.

\bibitem[Lenc and Vedaldi(2015)]{equivariance}
Karel Lenc and Andrea Vedaldi.
\newblock Understanding image representations by measuring their equivariance and equivalence.
\newblock In \emph{Proceedings of the IEEE/CVF Conference on Computer Vision and Pattern Recognition}, 2015.

\bibitem[Lin et~al.(2023)Lin, Gao, Tang, Takikawa, Zeng, Huang, Kreis, Fidler, Liu, and Lin]{magic3D}
Chen-Hsuan Lin, Jun Gao, Luming Tang, Towaki Takikawa, Xiaohui Zeng, Xun Huang, Karsten Kreis, Sanja Fidler, Ming-Yu Liu, and Tsung-Yi Lin.
\newblock Magic3d: High-resolution text-to-3d content creation.
\newblock In \emph{Proceedings of the IEEE/CVF Conference on Computer Vision and Pattern Recognition}, 2023.

\bibitem[Liu et~al.(2022)Liu, Bao, Sun, and Mei]{liu2022recent}
Wu Liu, Qian Bao, Yu Sun, and Tao Mei.
\newblock Recent advances of monocular 2d and 3d human pose estimation: A deep learning perspective.
\newblock \emph{ACM Computing Surveys}, 2022.

\bibitem[Liu et~al.(2015)Liu, Luo, Wang, and Tang]{celeba}
Ziwei Liu, Ping Luo, Xiaogang Wang, and Xiaoou Tang.
\newblock Deep learning face attributes in the wild.
\newblock In \emph{Proceedings of the IEEE International Conference on Computer Vision}, 2015.

\bibitem[Liu et~al.(2016)Liu, Luo, Qiu, Wang, and Tang]{DeepFashion}
Ziwei Liu, Ping Luo, Shi Qiu, Xiaogang Wang, and Xiaoou Tang.
\newblock Deepfashion: Powering robust clothes recognition and retrieval with rich annotations.
\newblock In \emph{Proceedings of the IEEE/CVF Conference on Computer Vision and Pattern Recognition}, 2016.

\bibitem[Lorenz et~al.(2019)Lorenz, Bereska, Milbich, and Ommer]{lorenz2019unsupervised}
Dominik Lorenz, Leonard Bereska, Timo Milbich, and Bjorn Ommer.
\newblock Unsupervised part-based disentangling of object shape and appearance.
\newblock In \emph{Proceedings of the IEEE/CVF Conference on Computer Vision and Pattern Recognition}, 2019.

\bibitem[Lowe(2004)]{SIFT}
David~G. Lowe.
\newblock {Distinctive Image Features from Scale-Invariant Keypoints}.
\newblock \emph{International Journal of Computer Vision}, 2004.

\bibitem[Luiten et~al.(2023)Luiten, Kopanas, Leibe, and Ramanan]{luiten2023dynamic}
Jonathon Luiten, Georgios Kopanas, Bastian Leibe, and Deva Ramanan.
\newblock Dynamic 3d gaussians: Tracking by persistent dynamic view synthesis.
\newblock \emph{arXiv Preprint}, 2023.

\bibitem[Luo et~al.(2023)Luo, Dunlap, Park, Holynski, and Darrell]{diffusionHyperfeatures}
Grace Luo, Lisa Dunlap, Dong~Huk Park, Aleksander Holynski, and Trevor Darrell.
\newblock Diffusion hyperfeatures: Searching through time and space for semantic correspondence.
\newblock \emph{Advances in Neural Information Processing Systems}, 2023.

\bibitem[Marullo et~al.(2023)Marullo, Tanzi, Piazzolla, and Vezzetti]{marullo20236d}
Giorgia Marullo, Leonardo Tanzi, Pietro Piazzolla, and Enrico Vezzetti.
\newblock 6d object position estimation from 2d images: a literature review.
\newblock \emph{Multimedia Tools and Applications}, 2023.

\bibitem[Metzer et~al.(2023)Metzer, Richardson, Patashnik, Giryes, and Cohen-Or]{latentNerf}
Gal Metzer, Elad Richardson, Or Patashnik, Raja Giryes, and Daniel Cohen-Or.
\newblock Latent-nerf for shape-guided generation of 3d shapes and textures.
\newblock In \emph{Proceedings of the IEEE/CVF Conference on Computer Vision and Pattern Recognition}, 2023.

\bibitem[Mokady et~al.(2022)Mokady, Hertz, Aberman, Pritch, and Cohen-Or]{prompt-to-prompt}
Ron Mokady, Amir Hertz, Kfir Aberman, Yael Pritch, and Daniel Cohen-Or.
\newblock Null-text inversion for editing real images using guided diffusion models.
\newblock \emph{Proceedings of the IEEE/CVF Conference on Computer Vision and Pattern Recognition}, 2022.

\bibitem[Nandy et~al.(2022)Nandy, Duan, and Kulik]{nandy2022audacity}
Aditya Nandy, Chenru Duan, and Heather~J Kulik.
\newblock Audacity of huge: overcoming challenges of data scarcity and data quality for machine learning in computational materials discovery.
\newblock \emph{Current Opinion in Chemical Engineering}, 2022.

\bibitem[Nichol and Dhariwal(2021)]{ddpm}
Alexander~Quinn Nichol and Prafulla Dhariwal.
\newblock Improved denoising diffusion probabilistic models.
\newblock In \emph{International Conference on Machine Learning}, 2021.

\bibitem[OpenAI(2023)]{gpt4}
OpenAI.
\newblock Gpt-4 technical report, 2023.

\bibitem[Papandreou et~al.(2018)Papandreou, Zhu, Chen, Gidaris, Tompson, and Murphy]{papandreou2018personlab}
George Papandreou, Tyler Zhu, Liang-Chieh Chen, Spyros Gidaris, Jonathan Tompson, and Kevin Murphy.
\newblock Personlab: Person pose estimation and instance segmentation with a bottom-up, part-based, geometric embedding model.
\newblock In \emph{Proceedings of the European Conference on Computer Vision}, 2018.

\bibitem[Poole et~al.(2022)Poole, Jain, Barron, and Mildenhall]{dreamFusion}
Ben Poole, Ajay Jain, Jonathan~T Barron, and Ben Mildenhall.
\newblock Dreamfusion: Text-to-3d using 2d diffusion.
\newblock \emph{International Conference on Learning Representations}, 2022.

\bibitem[Qu et~al.(2022)Qu, Liu, Liu, Wang, and Song]{qu2022towards}
Linhao Qu, Siyu Liu, Xiaoyu Liu, Manning Wang, and Zhijian Song.
\newblock Towards label-efficient automatic diagnosis and analysis: a comprehensive survey of advanced deep learning-based weakly-supervised, semi-supervised and self-supervised techniques in histopathological image analysis.
\newblock \emph{Physics in Medicine \& Biology}, 2022.

\bibitem[Ramesh et~al.(2022)Ramesh, Dhariwal, Nichol, Chu, and Chen]{dalle}
Aditya Ramesh, Prafulla Dhariwal, Alex Nichol, Casey Chu, and Mark Chen.
\newblock Hierarchical text-conditional image generation with clip latents.
\newblock \emph{arXiv Preprint}, 2022.

\bibitem[Rombach et~al.(2022)Rombach, Blattmann, Lorenz, Esser, and Ommer]{stableDiffusion}
Robin Rombach, Andreas Blattmann, Dominik Lorenz, Patrick Esser, and Bj{\"o}rn Ommer.
\newblock High-resolution image synthesis with latent diffusion models.
\newblock In \emph{Proceedings of the IEEE/CVF Conference on Computer Vision and Pattern Recognition}, 2022.

\bibitem[Russello et~al.(2022)Russello, van~der Tol, and Kootstra]{russello2022t}
Helena Russello, Rik van~der Tol, and Gert Kootstra.
\newblock T-leap: Occlusion-robust pose estimation of walking cows using temporal information.
\newblock \emph{Computers and Electronics in Agriculture}, 2022.

\bibitem[Saharia et~al.(2022)Saharia, Chan, Saxena, Li, Whang, Denton, Ghasemipour, Gontijo~Lopes, Karagol~Ayan, Salimans, et~al.]{imagen}
Chitwan Saharia, William Chan, Saurabh Saxena, Lala Li, Jay Whang, Emily~L Denton, Kamyar Ghasemipour, Raphael Gontijo~Lopes, Burcu Karagol~Ayan, Tim Salimans, et~al.
\newblock Photorealistic text-to-image diffusion models with deep language understanding.
\newblock \emph{Advances in Neural Information Processing Systems}, 2022.

\bibitem[Schmidtke et~al.(2021)Schmidtke, Vlontzos, Ellershaw, Lukens, Arichi, and Kainz]{schmidtke2021unsupervised}
Luca Schmidtke, Athanasios Vlontzos, Simon Ellershaw, Anna Lukens, Tomoki Arichi, and Bernhard Kainz.
\newblock Unsupervised human pose estimation through transforming shape templates.
\newblock In \emph{Proceedings of the IEEE/CVF Conference on Computer Vision and Pattern Recognition}, 2021.

\bibitem[Schuhmann et~al.(2022)Schuhmann, Beaumont, Vencu, Gordon, Wightman, Cherti, Coombes, Katta, Mullis, Wortsman, et~al.]{laion5b}
Christoph Schuhmann, Romain Beaumont, Richard Vencu, Cade Gordon, Ross Wightman, Mehdi Cherti, Theo Coombes, Aarush Katta, Clayton Mullis, Mitchell Wortsman, et~al.
\newblock Laion-5b: An open large-scale dataset for training next generation image-text models.
\newblock \emph{Advances in Neural Information Processing Systems}, 2022.

\bibitem[Siarohin et~al.(2019{\natexlab{a}})Siarohin, Lathuilière, Tulyakov, Ricci, and Sebe]{Siarohin_2019_CVPR}
Aliaksandr Siarohin, Stéphane Lathuilière, Sergey Tulyakov, Elisa Ricci, and Nicu Sebe.
\newblock Animating arbitrary objects via deep motion transfer.
\newblock In \emph{Conference on Computer Vision and Pattern Recognition}, 2019{\natexlab{a}}.

\bibitem[Siarohin et~al.(2019{\natexlab{b}})Siarohin, Lathuilière, Tulyakov, Ricci, and Sebe]{taichi}
Aliaksandr Siarohin, Stéphane Lathuilière, Sergey Tulyakov, Elisa Ricci, and Nicu Sebe.
\newblock First order motion model for image animation.
\newblock In \emph{Advances in Neural Information Processing Systems}, 2019{\natexlab{b}}.

\bibitem[Siarohin et~al.(2021)Siarohin, Roy, Lathuili{\`e}re, Tulyakov, Ricci, and Sebe]{siarohin2021motion}
Aliaksandr Siarohin, Subhankar Roy, St{\'e}phane Lathuili{\`e}re, Sergey Tulyakov, Elisa Ricci, and Nicu Sebe.
\newblock Motion-supervised co-part segmentation.
\newblock In \emph{International Conference on Pattern Recognition}, 2021.

\bibitem[Sun et~al.(2017)Sun, Shrivastava, Singh, and Gupta]{sun2017revisiting}
Chen Sun, Abhinav Shrivastava, Saurabh Singh, and Abhinav Gupta.
\newblock Revisiting unreasonable effectiveness of data in deep learning era.
\newblock In \emph{Proceedings of the IEEE International Conference on Computer Vision}, 2017.

\bibitem[Sun et~al.(2022)Sun, Ryou, Goldshmid, Weissbourd, Dabiri, Anderson, Kennedy, Yue, and Perona]{sun2022self}
Jennifer~J Sun, Serim Ryou, Roni~H Goldshmid, Brandon Weissbourd, John~O Dabiri, David~J Anderson, Ann Kennedy, Yisong Yue, and Pietro Perona.
\newblock Self-supervised keypoint discovery in behavioral videos.
\newblock In \emph{Proceedings of the IEEE/CVF Conference on Computer Vision and Pattern Recognition}, 2022.

\bibitem[Tang et~al.(2023)Tang, Jia, Wang, Phoo, and Hariharan]{emergentCorrespondences}
Luming Tang, Menglin Jia, Qianqian Wang, Cheng~Perng Phoo, and Bharath Hariharan.
\newblock Emergent correspondence from image diffusion.
\newblock \emph{Advances in Neural Information Processing Systems}, 2023.

\bibitem[Thewlis et~al.(2017)Thewlis, Bilen, and Vedaldi]{thewlis2017unsupervised}
James Thewlis, Hakan Bilen, and Andrea Vedaldi.
\newblock Unsupervised learning of object landmarks by factorized spatial embeddings.
\newblock In \emph{Proceedings of the IEEE International Conference on Computer Vision}, 2017.

\bibitem[Tian et~al.(2023)Tian, Aggarwal, Colaco, Kira, and Gonzalez-Franco]{diffuseAttendSegment}
Junjiao Tian, Lavisha Aggarwal, Andrea Colaco, Zsolt Kira, and Mar Gonzalez-Franco.
\newblock Diffuse, attend, and segment: Unsupervised zero-shot segmentation using stable diffusion.
\newblock \emph{arXiv Preprint}, 2023.

\bibitem[Touvron et~al.(2023)Touvron, Lavril, Izacard, Martinet, Lachaux, Lacroix, Rozi{\`e}re, Goyal, Hambro, Azhar, et~al.]{llama}
Hugo Touvron, Thibaut Lavril, Gautier Izacard, Xavier Martinet, Marie-Anne Lachaux, Timoth{\'e}e Lacroix, Baptiste Rozi{\`e}re, Naman Goyal, Eric Hambro, Faisal Azhar, et~al.
\newblock Llama: Open and efficient foundation language models.
\newblock \emph{arXiv Preprint}, 2023.

\bibitem[Wah et~al.(2011)Wah, Branson, Welinder, Perona, and Belongie]{cub}
Catherine Wah, Steve Branson, Peter Welinder, Pietro Perona, and Serge Belongie.
\newblock {The Caltech-UCSD Birds-200-2011 Dataset}.
\newblock Technical report, California Institute of Technology, 2011.

\bibitem[Wang et~al.(2023{\natexlab{a}})Wang, Li, Zhang, Xu, Zhou, Yu, Sheng, and Xu]{DiffSegmenter}
Jinglong Wang, Xiawei Li, Jing Zhang, Qingyuan Xu, Qin Zhou, Qian Yu, Lu Sheng, and Dong Xu.
\newblock Diffusion model is secretly a training-free open vocabulary semantic segmenter.
\newblock \emph{arXiv Preprint}, 2023{\natexlab{a}}.

\bibitem[Wang et~al.(2023{\natexlab{b}})Wang, Chang, Cai, Li, Hariharan, Holynski, and Snavely]{wang2023tracking}
Qianqian Wang, Yen-Yu Chang, Ruojin Cai, Zhengqi Li, Bharath Hariharan, Aleksander Holynski, and Noah Snavely.
\newblock Tracking everything everywhere all at once.
\newblock \emph{International Conference on Computer Vision}, 2023{\natexlab{b}}.

\bibitem[Wu et~al.(2023)Wu, Zhao, Shou, Zhou, and Shen]{DiffuMask}
Weijia Wu, Yuzhong Zhao, Mike~Zheng Shou, Hong Zhou, and Chunhua Shen.
\newblock Diffumask: Synthesizing images with pixel-level annotations for semantic segmentation using diffusion models.
\newblock \emph{arXiv Preprint}, 2023.

\bibitem[Wu and Ji(2019)]{wu2019facial}
Yue Wu and Qiang Ji.
\newblock Facial landmark detection: A literature survey.
\newblock \emph{International Journal of Computer Vision}, 2019.

\bibitem[Xiao et~al.(2023)Xiao, Yang, Zhou, and Zhang]{textToMask}
Changming Xiao, Qi Yang, Feng Zhou, and Changshui Zhang.
\newblock From text to mask: Localizing entities using the attention of text-to-image diffusion models.
\newblock \emph{arXiv Preprint}, 2023.

\bibitem[Xu et~al.(2023)Xu, Liu, Vahdat, Byeon, Wang, and De~Mello]{odise}
Jiarui Xu, Sifei Liu, Arash Vahdat, Wonmin Byeon, Xiaolong Wang, and Shalini De~Mello.
\newblock Open-vocabulary panoptic segmentation with text-to-image diffusion models.
\newblock In \emph{Proceedings of the IEEE/CVF Conference on Computer Vision and Pattern Recognition}, 2023.

\bibitem[Xu et~al.(2022)Xu, Jin, Liu, Qian, Ouyang, Luo, and Wang]{xu2022zoomnas}
Lumin Xu, Sheng Jin, Wentao Liu, Chen Qian, Wanli Ouyang, Ping Luo, and Xiaogang Wang.
\newblock Zoomnas: searching for whole-body human pose estimation in the wild.
\newblock \emph{IEEE Transactions on Pattern Analysis and Machine Intelligence}, 2022.

\bibitem[Zhang et~al.(2023)Zhang, Herrmann, Hur, Cabrera, Jampani, Sun, and Yang]{TaleOFTwoFeats}
Junyi Zhang, Charles Herrmann, Junhwa Hur, Luisa~Polania Cabrera, Varun Jampani, Deqing Sun, and Ming-Hsuan Yang.
\newblock A tale of two features: Stable diffusion complements dino for zero-shot semantic correspondence.
\newblock \emph{Advances in Neural Information Processing Systems}, 2023.

\bibitem[Zhang et~al.(2018)Zhang, Guo, Jin, Luo, He, and Lee]{zhang2018unsupervised}
Yuting Zhang, Yijie Guo, Yixin Jin, Yijun Luo, Zhiyuan He, and Honglak Lee.
\newblock Unsupervised discovery of object landmarks as structural representations.
\newblock In \emph{Proceedings of the IEEE/CVF Conference on Computer Vision and Pattern Recognition}, 2018.

\bibitem[Zhang et~al.(2022)Zhang, Liang, Zou, Li, Sun, and Wang]{zhang2022self}
Yanping Zhang, Qiaokang Liang, Kunlin Zou, Zhengwei Li, Wei Sun, and Yaonan Wang.
\newblock Self-supervised part segmentation via motion imitation.
\newblock \emph{Image and Vision Computing}, 2022.

\bibitem[Zheng et~al.(2023)Zheng, Wu, Chen, Yang, Zhu, Shen, Kehtarnavaz, and Shah]{zheng2023deep}
Ce Zheng, Wenhan Wu, Chen Chen, Taojiannan Yang, Sijie Zhu, Ju Shen, Nasser Kehtarnavaz, and Mubarak Shah.
\newblock Deep learning-based human pose estimation: A survey.
\newblock \emph{ACM Computing Surveys}, 2023.

\end{thebibliography}
}

\end{document}